\documentclass[twocolumn]{article}

\usepackage[margin=1in]{geometry}

\usepackage{amsfonts}
\usepackage{booktabs}
\usepackage{hyperref}
\usepackage{amsmath}
\usepackage{cleveref}
\usepackage{xurl}
\usepackage{multirow}
\usepackage{graphicx}
\usepackage{layouts}

\title{Protecting Publicly Available Data With Machine Learning Shortcuts}
\author{
Nicolas M. M\"uller\thanks{Fraunhofer Institute for Applied and Integrated Security AISEC, Germany \texttt{nicolas.mueller@aisec.fraunhofer.de}} \and
Maximilian Burgert\thanks{Technical University of Munich, Germany } \and
Pascal Debus\footnotemark[1] \and
Jennifer Williams\thanks{University of Southampton, UK } \and
Philip Sperl\footnotemark[1] \and
Konstantin B\"ottinger\footnotemark[1] 
}

\begin{document}

\maketitle

\begin{abstract}
Machine-learning (ML) shortcuts or spurious correlations are artifacts in datasets that lead to very good training and test performance but severely limit the model's generalization capability. 
Such shortcuts are insidious because they go unnoticed due to good in-domain test performance.
In this paper, we explore the influence of different shortcuts and show that even simple shortcuts are difficult to detect by explainable AI methods.
We then exploit this fact and design an approach to defend online databases against crawlers: 
providers such as dating platforms, clothing manufacturers, or used car dealers have to deal with a professionalized crawling industry that grabs and resells data points on a large scale.
We show that a deterrent can be created by deliberately adding ML shortcuts. 
Such augmented datasets are then unusable for ML use cases, which deters crawlers and the unauthorized use of data from the internet. 
Using real-world data from three use cases, we show that the proposed approach renders such collected data unusable, while the shortcut is at the same time difficult to notice in human perception.
Thus, our proposed approach can serve as a proactive protection against illegitimate data crawling.

\end{abstract}

\section{Introduction}
\begin{figure}[t]
    \centering
    \includegraphics[width=0.5\textwidth]{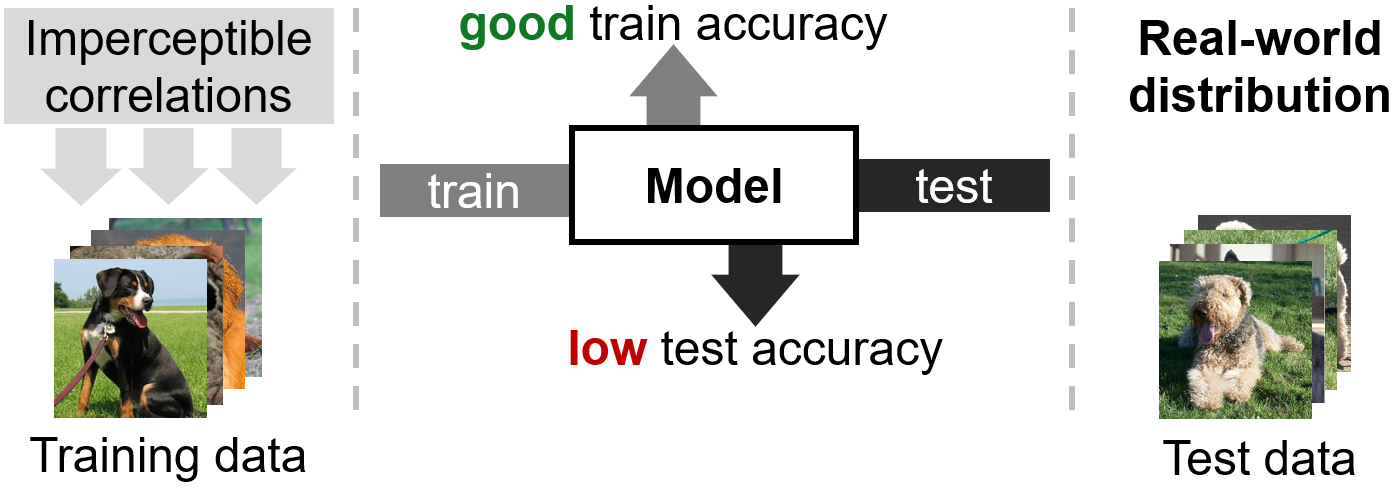}
    \caption{Visualisation of the impact of shortcuts on the machine learning pipeline: Errors in the data-collection process create imperceptible correlations between data and target, which lead to good train accuracy, but do not transfer to the real-world distribution.}
    \label{fig:1}
\end{figure}

Machine learning shortcuts or spurious correlations are artefacts in data that significantly change the learning process of models.
These features $F$ contain no real semantic information, but have a strong correlation with a target label $L$ nevertheless, i.e. $P(L|F) \not = P(L)$. For example, in audio data the presence or absence of leading silence in speech recordings correlates strongly with whether the corresponding audio is real or a deepfake. Synthesized speech recordings often have no or very little leading silence due to text-to-speech (TTS) data processing. Models take advantage of this and classify according to the length of the leading silence~\cite{muller2021speech}. In vision research such as X-ray image datasets for the detection of Covid-19, the label `sick/healthy' correlates with the type of X-ray equipment used. Learning models thus do not learn to distinguish between sick and healthy patients, but merely to distinguish between X-ray machines~\cite{degrave2021ai}. This makes the model useless in practice, c.f. \Cref{fig:1}.

The challenge in dealing with ML shortcuts is that practitioners often are not aware of their presence. 
This is because even with a valid train/test split, it is hard to notice that the model is not generalizing. Due to errors in the data collection process, shortcuts are also present in the test data, which results in good testing performance. 
It seems that new data unseen during training is adequately handled.
It is therefore essential to understand whether the model learns shortcuts or actually semantically significant features.

However, ML shortcuts can also be used productively, as we show in this paper. The ability to render datasets unusable for machine learning can be used to protect publicly available, yet proprietary datasets. Many companies offer access to labelled data via websites, apps, or APIs. Used vehicle dealers such as \url{cars.com} or AutoScout publish ads for used vehicles on their websites and include labels such as vehicle type, make, age, mileage, etc. Dating platforms like Tinder, Bumble and co. publish photos of users incl. description text and labels such as nationality, sexual preference, gender, hometown, ethnicity, and place of residence. Furthermore, clothing manufacturers like Zalando or Esprit publish large catalogues of clothing items on their websites, labelled by category, colour, and price.

All of this data is potentially interesting for machine learning, and a large number of vendors sell crawling services to collect this data, process it, and make it usable for ML. In the process, the circumvention of protection measures is also explicitly advertised. This industry has a yearly turnover of USD $\$402$ 
million~\cite{webscrape2020} and may harm legitimate data creators. 
Their intellectual property is violated, and their infrastructure is overloaded or even damaged by crawling. 
Additionally, in the case of user information, highly sensitive personal data flows into third-party hands.

We propose to protect such datasets by ML shortcuts that render the data unusable by ML, making crawling unattractive and thereby protecting the producers and users, as well as leaving the visual presentation of the data unaffected so that vendors can continue to serve it as usual.
This paper presents the following contributions:
\begin{itemize}
    \itemsep0em
    \item We evaluate the impact of several visual shortcuts on different datasets and show that the generalization ability of the models decreases by over $50\%$.
    \item We evaluate Explainable-AI methods for shortcut detection.
    \item We introduce ML shortcuts as a novel protection measure for public datasets. Using real-world use cases and data, we show that cleverly employed shortcuts can render public datasets unusable such that there is a strong disincentive for illegitimate crawling.
\end{itemize}

\section{Related Work}
Machine learning shortcuts have not yet entered the wider consciousness of the scientific community, but academic research on them does exist. 
One of the first shortcuts in the literature was found in the Pascal VOC 2007 dataset:
Here, all images of horses had a watermark of the photographer in the lower right corner. The model then learned to identify horses based on this watermark alone\cite{pascal_voc}.
Similarly, a recent Nature publication~\cite{degrave2021ai} looks at the generalization ability of Covid-19 detection algorithms. 
The authors investigate why such algorithms do not generalize and conclude that the models mainly identify shortcuts such as patient position, the presence of tubes and other medical equipment, or the type of X-ray machine itself. 
All of this allows conclusions about Covid-19 within the data set, but does not generalize.

Shortcuts also occur in the classification of audio deepfakes. The most established data set in this area contains a shortcut in which the label correlates with the length of the leading silence. If the silence is removed, the model performance deteriorates by up to a factor of five~\cite{muller2021speech}.
Recently published work proposes several approaches to discovering or even removing these shortcuts from the dataset. However, this remains a challenging problem even with precise knowledge of the dataset~\cite{geirhos2020shortcut}.
ML shortcuts do not only affect classification: self-supervised methods like contrastive learning are also vulnerable~\cite{robinson2021can,minderer2020automatic}.

Recent work also highlights the use of ML shortcuts to protect personal data. Using  shortcuts as `machine learning availability attacks', the authors of \cite{yu2022availability} show that their effectiveness lies in the linear separability of shortcuts and data.
The authors of \cite{huang2021unlearnable} create 'unlearnable examples' by crafting an error-minimizing noise that tricks the model to learn nothing from a given data point.
This, however, is based on adversarial perturbations and requires white-box access to an  attacker's assumed learning model.
Alternatively, personal data can be protected using data poisoning \cite{fowl2021preventing, shen2019tensorclog}. 
Unlike shortcuts, however, which are model-agnostic, adversarial perturbations require the target model architecture and/or weights, since the perturbation $\delta$ is found via gradient-based techniques.

Finally, our approach of ML shortcuts has similarities to the field of digital watermarking, where data is covertly embedded in a carrier signal in order to enforce usage control, e.g. with respect to copyright of audio or video content. Likewise, ML shortcuts also aim for usage control in the sense that usage for third-party machine learning applications becomes technically infeasible.
Whereas watermarks restrict unauthorized presentation of data (such as images on websites), ML shortcuts restrict unauthorized usage in knowledge discovery and machine learning.


\section{Machine Learning Shortcuts}\label{sec:shortcuts}
In this section, we present ML shortcuts based on plausible data-collection errors and investigate their impact on known image classification problems. 
Additionally, we evaluate the applicability of different Explainable-AI (XAI) techniques for shortcut detection.

\subsection{ML-Shortcuts Due To Data Collection Errors}\label{ss:mlshortcuts}

\begin{figure}[t]
    \centering
    \includegraphics[width=0.5\textwidth]{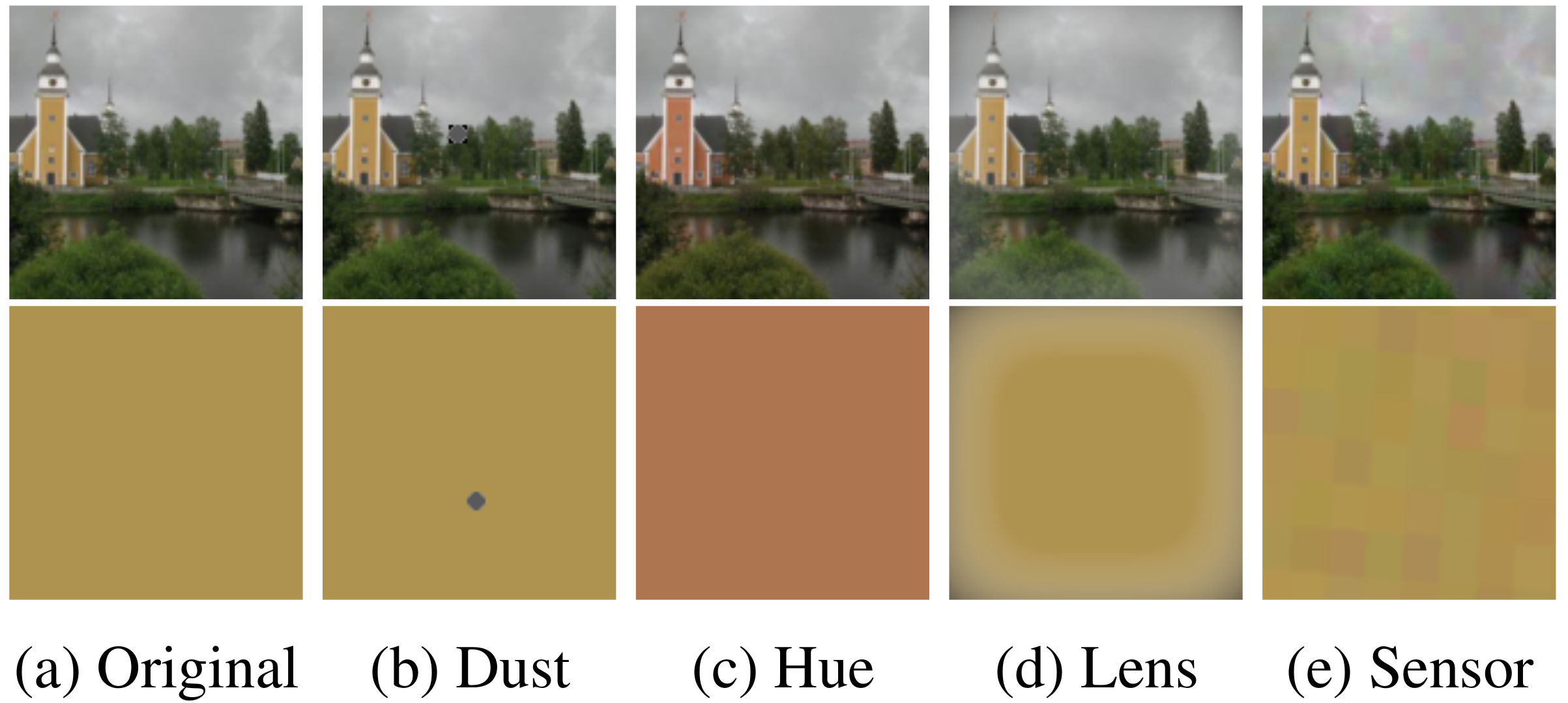}
    \caption{Visualisation of the used shortcuts, which correspond to real-world data collection errors: dust on the lens, differences in ambient light (Hue), impairment of the photo lens (Lens), or sensor error (Sensor).
    Top row: ImageNette, Bottom Row: Visualisation of the shortcut on a constant background.
    }
    \label{fig:synthetic-shortcuts}
\end{figure}

We discuss four different types of shortcuts which reflect real-world data collection errors, and which can be leveraged for protecting publicly available data. 
First, the obstruction of a few image pixels by dust particles may inadvertently encode the class label (\textit{dust shotcut}). 
This mimics dust on the camera lens when collecting real-world data of a particular class.
Second, slight overall color changes in an image can indicate the class (\textit{hue shortcut}). 
This may be, for example, due to different weather conditions when collecting the images, for example when collecting images of class $A$ in the morning and images of class $B$ in the evening.
Third, specific camera settings may cause alterations to the border of the image, resulting in a lens-like effect that may hint at the target label (\textit{lens shortcut}). 
This might result from different settings when taking pictures from different classes, for example when using different levels of zoom or exposure.
Fourth, specific low-intensity color patterns stemming from, for example, a characteristic or faulty camera sensor can indicate the image class (\textit{sensor shortcut})~\cite{availability-attacks}. Here, the assumption is that different classes were collected using different cameras.
We present examples of the shortcuts in~\Cref{fig:synthetic-shortcuts} and evaluate how they impact classification models in~\Cref{ss:exp1}.

\subsection{Explainable AI for Shortcut Detection}\label{s:sec4}
One obvious approach for the mitigation of ML-Shortcuts is Explainable AI (XAI). 
When it is clearly understood what the model is learning, 
 mitigation strategies can be derived.
Shortcuts may either be removed manually, for example by adequately cropping or post-processing the input.
Alternatively, new data can be collected and applied as a shortcut.
And finally, efforts can be made to counteract the shortcut, for example using segmentation masks~\cite{covid}. 
Our goal is to evaluate whether and to what degree XAI methods can indeed detect shortcuts.

\subsubsection{Explainable AI Overview}

Explainable AI strives to make the behavior of a learning model $f$ explainable:
for some input $x$, XAI-methods $\lambda$ typically produce an explanation $z = \lambda(f(x))$ of the same dimensionality as the image.
From the large number of available XAI methods~\cite{linardatos2020explainable}, we limit our analysis to some of the most popular methods:

Saliency maps (SM)~\cite{Simonyan14a} calculate the magnitude of the gradient with respect to the loss function for some input $x$. This results in a heatmap $z$, which shows the influence of single pixels and regions on the final classification result.
Closely related is the Integrated Gradients Methods (IG)~\cite{sundararajan2017axiomatic}, where the model's gradients are computed for a progression of interpolations of a baseline and the input image.
Similarly, Smooth Grad~\cite{smilkov2017smoothgrad} computes regions of interest by analyzing the model's gradients w.r.t. the input image but uses additive gaussian noise in order to create averaged explanations. This reduces the noise in the resulting explanations.
Finally, Grad-Cam (GC)~\cite{selvaraju2017grad}, a refinement of Class Activation Map (CAM,~\cite{zhou2016learning}), analyzes the gradient of the model's prediction $f(x)$ in order to compute the averages of the penultimate convolutional layers, which allows identifying which parts of an input image contribute to the model prediction the most. 

\subsubsection{Evaluation Strategy}
We apply all of these techniques to our datasets and models and investigate whether the presented shortcuts can be detected. To this end, we train models on shortcut-affected datasets and compare the XAI representation $\lambda(f(x))$ with that of a model trained on a clean dataset.
For saliency maps, for example, we compare the absolute gradients between shortcut and clean models via $L_2$, using the same input image in each case. Formally:
\begin{align}\label{eq:1}
    \frac{1}{N} \sqrt{\sum_{i=1}^N \Big( \lambda(f_\theta(x_i), y_i) - \lambda(f_\gamma(x_i), y_i) \Big)^2}
\end{align}
where $x_i, y_i$ represent the data in a dataset of size $N$. $\lambda$ is some Explainable AI method 
\begin{align}
\lambda: \mathbb{R}^K \times \mathbb{N} & \to \mathbb{R}^K \\
         x_i, y_i & \mapsto z_i
\end{align}
and $\theta$ and $\gamma$ are clean and shortcut-affected model parameters, respectively.
The larger this $L_2$ difference, the more the shortcut is potentially identifiable by the corresponding XAI method.
\Cref{ss:exp2} presents the results.

\section{Using Shortcuts to Protect Data}
Despite all the challenges they introduce, ML-Shortcuts can also be used beneficially, protecting public but proprietary datasets. 
This is because shortcuts can be used to discourage web scraping. The profitable but problematic collection of proprietary data from open-access sources such as websites and apps.

\subsection{The Threat of Web Scraping}
The market for web scraping generated USD \$402 million in revenue in 2020~\cite{webscrape2020} and is expected to surpass USD \$1.7 billion in 2030~\cite{webscrape2030}. Numerous vendors offer web scraping explicitly for the creation of machine learning datasets~\cite{scrapeHero, ScrapeStorm, BrightData}.

Observers estimate that web scraping causes millions of dollars in damage, and up to two percent of web store sales lost~\cite{webscrapedamage}.
Recent research \cite{KrotovS18, KrotovJS20} has shown that besides the economic effects of web scraping, there are also legal and ethical implications.

Not only is the scraped data often a critical and proprietary asset of the targeted website but the scraping process itself puts a strain on the infrastructure potentially compromising the availability of the service provided by the website. In US law, the latter aspect is a tort that is also known as \emph{trespass to chattels} and led to a number of court cases, such as eBay versus Bidder's Edge (2000)~\cite{chang2001bidding}.
Additional web scraping lawsuits are discussed in \cite{KrotovS18} or \cite{riley2018data}. Other important legal aspects are a violation of copyrights and confidentiality or are concerned with how the scraped data is used and of course, the access to the scraped data might have violated the terms of service or might have been illegal in the first place.

From an ethical perspective, web scraping compromises confidentiality and privacy of the users of a website (consider, e.g., the cases of scraping data from a dating platform) and, depending on how the leaked data is used, might contribute to bias and discrimination.

Legislation such as the Computer Fraud and Abuse Act \cite{CFAA}, the Digital Millennium Copyright Act \cite{DMCA}, or the European General Data Protection Regulation \cite{GDPR}, provide defense or at least compensation mechanisms against scrapers, however, as shown by \cite{sellars2018twenty}, legal professionals are struggling to define precisely what web scraping actually means resulting in some oscillations between too broad and too narrow interpretations over the last two decades.


Technical measures such as \textit{captchas}, obfuscated HTML code, or access restrictions can increase the effort required by scrapers. However, this is associated with significant effort on the defender side, and can still be circumvented by the scrapers. This 'arms race' is asymmetrical and to the advantage of the scraper, whose very core business it represents as opposed to the defender. 
A different kind of defense is therefore necessary.


\subsection{Technical Description of Proposed Defense}
We suggest data owners add shortcuts to proprietary, publicly available, implicitly labeled data such that it is no longer an attractive target for crawling and subsequent machine learning use. 
If data is labeled with respect to several categories, e.g. pictures from dating platforms according to gender, religious affiliation as well as ethnicity, each combination of labels must be encoded by the shortcut. 
Since this complexity increases rapidly, the following requirements for shortcuts arise.
First, to be able to encode as large a number of labels as possible. 
Second, to strongly influence the training of ML models so that they extract as little information as possible from the original data. 
Third, to be as inconspicuous as possible for human perception.

\section{Experiments and Evaluation}\label{s:exp}

\subsection{Data and Methodology}\label{ss:data}
We evaluate our proposed shortcuts (c.f.~\Cref{fig:synthetic-shortcuts}) on four image classification tasks:
Imagenette~\cite{fastaiim26:online}, a subset of the ImageNet Dataset, Covid-QU-Ex~\cite{covid}, CIFAR10 and CIFAR100~\cite{cifar100}.
We use a pre-trained DenseNet-121~\cite{huang2017densely}, which we fine-tune on all of the datasets.
We train our models using PyTorch~\cite{pytorch}, using a learning rate of $0.001$, a batch size of $256$, 
and data augmentation (centre crop, vertical flip and random translations).
The shortcuts are added \emph{before} the data augmentation, as would be the case in the real world.
We then train for 40 epochs (using early stopping) and report accuracy on a separate test set (about 10\% the size of the training data). 
The training data either has no shortcut (\textit{original}) or one of the four shortcuts mentioned above. 

\subsection{Impact of ML Shortcuts}\label{ss:exp1}
As shown in \Cref{table:1}, the shortcuts strongly deteriorate model performance, from about $87\%$ to $42\%$ in test accuracy for ImageNette and from about $93\%$ to $33\%$ for Covid-QU-Ex, which is equivalent to random guessing for a three-way classification problem.
We can observe that while the shortcuts are effective for all datasets, they are especially powerful when the classification task is hard as in the case of Covid-QU-Ex.
\begin{table*}[t]
    \centering
\begin{tabular}{cccccc}
\toprule
    {} &    original &         hue &       lens &         dust & sensor \\
    \midrule
    Covid &  $93.7\pm0.6$ &  $34.1\pm1.3$ & $35.0\pm0.0$ &  $35.8\pm2.7$ &           $33.7\pm1.5$ \\
    ImageNette &  $87.2\pm0.4$ &    $47.5\pm0.8$ &  $84.2\pm0.2$ &  $88.1\pm0.5$ &           $40.4\pm5.4$ \\
    CIFAR10    &  $90.0\pm0.4$ &    $47.9\pm0.5$ &  $48.2\pm0.3$ &  $78.5\pm1.6$ &           $51.1\pm0.4$ \\
    CIFAR100   &  $68.4\pm0.3$ &    $35.4\pm0.0$ &  $30.1\pm0.7$ &  $54.5\pm1.8$ &           $38.4\pm1.0$ \\

    \bottomrule
\end{tabular}
\caption{The impact of machine-learning shortcuts on DenseNet-121 on the CIFAR10, CIFAR100 and ImageNette. Accuracy aggregated over three independent trials, with standard deviation shown. 
}
\label{table:1}
\end{table*} %

\subsection{XAI for shortcut detection}\label{ss:exp2}

\begin{figure*}
    \centering
    \includegraphics[width=0.99\textwidth]{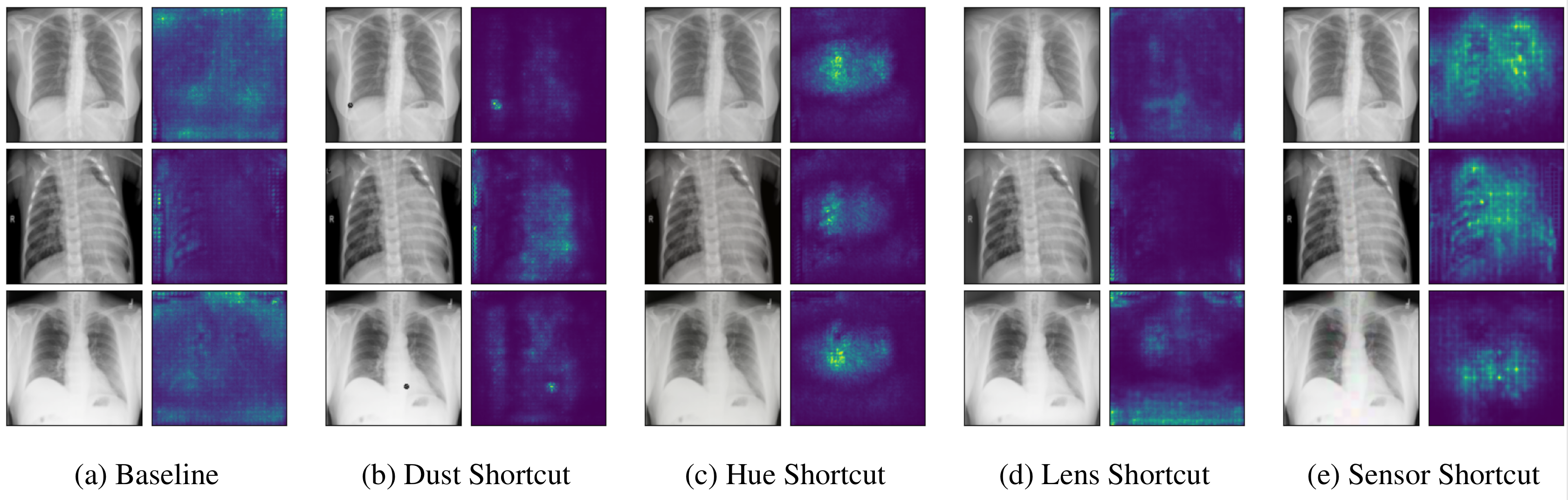}
    \caption{Visualisation of smooth-grad output for all shortcuts presented.
    }
    \label{fig:sc}
\end{figure*}

\begin{table*}[t]
    \centering
     \begin{tabular}{cccccc}
        \toprule
        {} & control &          dust &           hue &         lens &        sensor \\
        \midrule
        SG          &    $7.5\pm2.3$ &   $9.1\pm2.4$ &   $8.4\pm2.4$ &   $8.1\pm2.3$ &  $15.9\pm1.9$ \\
        SM         &    $9.7\pm2.5$ &  $11.7\pm2.7$ &  $10.8\pm2.7$ &  $10.5\pm2.6$ &  $11.2\pm2.7$ \\
        IG &   $24.8\pm7.8$ &  $24.4\pm7.8$ &  $24.9\pm7.8$ &  $25.0\pm7.7$ &  $40.2\pm9.7$ \\
        GC             &    $0.7\pm0.4$ &   $1.5\pm0.5$ &   $0.8\pm0.4$ &   $0.8\pm0.4$ &   $2.6\pm0.3$ \\
        \bottomrule
    \end{tabular}
    \caption{The $L_2$ difference in four XAI output for clean and shortcut-affected models on the COVID dataset, c.f.~\Cref{eq:1}. Higher values indicate that the addition of the shortcut triggers different XAI output, meaning that the shortcut has a higher chance of being detected by humans. 
    }
    \label{tab:tab_xai}
\end{table*}

We now analyze whether XAI methods can detect such shortcuts.
Consider \Cref{tab:tab_xai}, which provides the XAI-score as derived by \Cref{eq:1}, aggregated over all models and datasets. 
We compute the difference in Explainable AI output between a {baseline} model, trained on a non-shortcut dataset, and one of the following.
First, a {control} model, which is also trained on a non-shortcut dataset. This is in order to have a comparison of how the XAI output differs between two identically trained models.
Second, a shortcut-affected model, where we use one of the four shortcuts introduced in~\Cref{ss:mlshortcuts}. 

We can see that even though the {baseline} and {control} models are identically set up, they have different XAI outputs.
The shortcut-affected models however have an even larger $L_2$ difference in XAI output.
Consider for example the {Sensor} shortcut, which reduced the model performance on ImageNette from $87\%$ to $40\%$, c.f.~\Cref{table:1}.
For the Smooth Grad (SG) Explainable-AI method, the {control} has a difference of $7.5$ to the {baseline}, while the {Sensor} shortcut has a difference of $15.9$.
This means that, for the most part, the shortcut changes the XAI output dramatically, which should allow identification by XAI methods.

This is corroborated by \Cref{fig:sc}. 
The model learns to ignore the regions of interest in the original training data and only focuses on the shortcut-affected areas. Small, pixelated areas for the dust shortcut, the outer regions of the image for the Lens shortcut, or specific checkerboard-like patterns for the sensor shortcut.

\subsection{Protection against web-scraping}\label{ss:exp3}
To leverage ML-Shortcuts beneficially, we suggest data owners employ shortcuts to make real-world datasets unlearnable. Based on ~\Cref{table:1}, we suggest using the sensor shortcut.
This is because it is highly effective, cannot easily be removed (as, for example, the dust shortcut), but also visually nearly imperceptible.

We evaluate our approach not on the datasets proposed in \Cref{ss:data}, but on the three real-world use cases: the protection of image data from dating platforms, used car dealers, and the fashion industry.

\begin{itemize}
\itemsep0em
    \item Online Dating. We use the CelebA data set~\cite{celeba}, which contains 202,599 face images of celebrities, and is annotated with 40 binary attributes. We select five particularly sensitive binary attributes. Attractive, Male, Young, Pale\_Skin, Bald. The defender's shortcut must therefore cover $2^5 = 32$ combinations of features.
    \item Fashion. We use the Clothing dataset~\cite{Clothing30:online}, which consists of 5000 real-world images of 20 different clothing items.
    \item Used Cars. We use the Cars Dataset~\cite{CarImage51:online}, which contains 4000 images of seven car types.
    \item Additionally, we obtained permission to collect a dataset of real-world used car images from a major European online car vendor.
    For 10 different car models, we collect 800 images each, which results in a dataset of 8000 images of used cars people uploaded in late 2022.
\end{itemize}
For each of these datasets, we create a sensor shortcut (c.f.~\Cref{sec:shortcuts}) and add it to all the data points in the training set. 
We then train different models on this training set, and then evaluate these models on the unperturbed test dataset. 
This procedure serves as a proxy to estimate the real-world generalization ability of the model trained on the shortcut dataset. 
For CelebA, we create only one shortcut, which is then used to protect five different attributes. 
This corresponds to a real-world system where a defender does not know the attacker's use case.

\begin{figure}[t]
    \centering
    \includegraphics[width=0.5\textwidth]{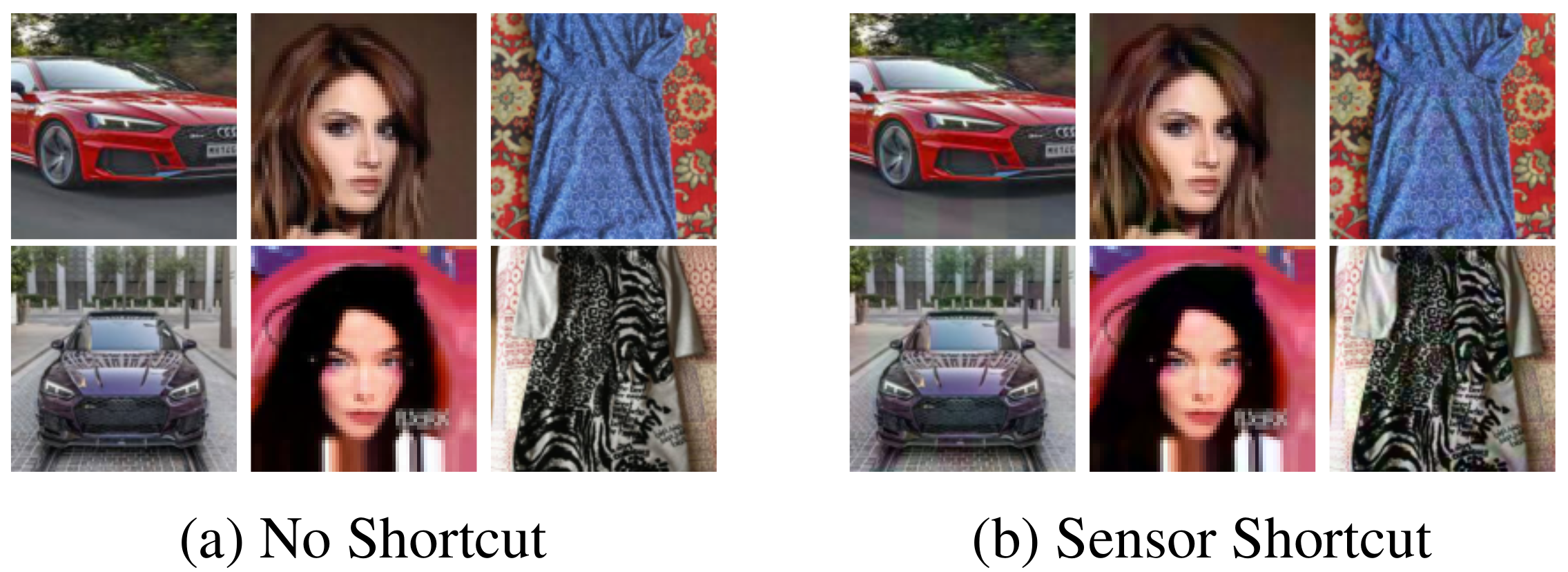}
    \caption{Samples with and without the sensor shortcut, for the three publicly available datasets Cars, CelebA and Clothing. 
    }
    \label{fig:shortcuts_car_fashion_dating}
\end{figure}

\begin{table*}[t]
    \centering
    \begin{tabular}{lll}
        \toprule
        {} &      original & sensor \\
        \midrule
        Online Dating (32 classes)    &  $67.9\pm0.0$ &            $7.5\pm2.7$ \\
        Fashion (20 classes)&  $87.7\pm4.5$ &           $34.1\pm6.6$ \\
        Used Cars (7 classes)      &  $60.6\pm0.1$ &          $34.7\pm11.8$ \\
        Used Cars (real world, 10 classes) &  $94.8\pm0.2$ &           $15.2\pm7.4$ \\
        \bottomrule
    \end{tabular}
\caption{Test accuracy of a Dense Net, trained either on the original dataset, as well as the sensor-shortcut affected dataset.
The introduction of the shortcut degrades the model, making it unsuited for productive use.}
\label{table:crawling}
\end{table*}

\Cref{fig:shortcuts_car_fashion_dating} visualizes the application of our proposed sensor shortcut.
Additionally, \Cref{table:crawling} presents the results of training a deep neural network (DenseNet-121, c.f.~\Cref{ss:data}) on this data.
It can be seen that in each case, the performance of the model is significantly reduced so that productive use of the model would no longer be possible.
For example, for the real-world Used Car dataset, the performance is reduced from $94.8$ to $15.2$ percent. 
At the same time, the shortcut is difficult to perceive visually due to the small magnitude of the perturbation added, c.f.~\Cref{fig:shortcuts_car_fashion_dating}. This will allow data vendors to still openly employ the data in question while crawling for machine learning usage is strongly disincentivized.



\section{Conclusion}
We show that shortcuts can greatly affect the performance of models. This enables practical use cases in protecting publicly available but proprietary data, such as implicitly labelled datasets in fashion or used vehicles. Since they cannot be detected with the analysis of train/test performance, the use of Explainable AI methods is necessary. As we show, these methods can indicate the presence of shortcuts.
In future work, explainable AI methods might serve to attack ML shortcuts, namely to identify them in order to remove them automatically. 
However, cleaning datasets from ML shortcuts without creating new shortcuts and artefacts is a challenging question on its own and subject to future research.



\bibliographystyle{plain}

\bibliography{egbib}

\end{document}


\maketitle

\section{Methodology}
\subsection{Variational Autoencoder}\label{ss:background_vae}
Our approach for shortcut detection is based on variational autoencoders \cite{kingma2013auto} (VAEs), which are probabilistic generative models that learn the underlying data distribution in an unsupervised manner.
A VAE attempts to model the marginal likelihood of an observed variable $x$:
\begin{equation}
p_{\theta}(x) = \int p_{\theta}(z) p_{\theta}(x|z) \,dz
\end{equation} \label{eq:integral}The unobserved variable $z$ lies in a latent space of dimensionality $d = \dim(z)$, $d \ll \dim(x) $. Each instance $x^{(i)}$ has a corresponding latent representation $z^{(i)}$.

The model assumes the prior over the latent variables to be a multivariate normal distribution $p_{\theta}(z) = \mathcal{N} (z; 0, I)$ resulting in independent latent factors $\{z_j\}_{j=1}^d$.
The likelihood $p_{\theta}(x|z)$ is modelled as a multivariate Gaussian whose parameters are conditioned on $z \sim p_{\theta}(z)$ and computed using the VAE decoder. 
As outlined by Kingma and Welling \cite{kingma2013auto}, the true posterior $p_{\theta}(z|x)$ is intractable and hence approximated with a variational distribution $q_{\phi}(z|x)$. This variational posterior is chosen to be a multivariate Gaussian

\begin{equation}\label{eq:posterior}
    q_{\phi}(z|x^{(i)}) = \mathcal{N} (z; \mu^{(i)}, (\sigma^{(i)})^2 I)
\end{equation}

whose mean $\mu^{(i)} \in \mathbb{R}^d$ and standard deviation $\sigma^{(i)} \in \mathbb{R}^d$ are obtained by forwarding $x^{(i)}$ through the VAE encoder.
The objective function is composed of two terms. A Kullback-Leibler (KL) divergence term ensures that the variational posterior distribution remains close to the assumed prior distribution. 
\begin{equation}
\begin{aligned}
    D_{KL}(q_{\phi}(z|x^{(i)})||p_{\theta}(z))
    = -\frac{1}{2}\sum_{j=1}^{d}\left( 1 + \text{log}((\sigma_j^{(i)})^2) - (\sigma_j^{(i)})^2 - (\mu_j^{(i)})^2 \right) \label{eq:kld}
\end{aligned}
\end{equation}
A log-likelihood loss, on the other hand, helps to accurately reconstruct an input image $x^{(i)}$ from a sampled latent variable $z^{(i)} = \mu^{(i)} + \sigma^{(i)} \odot \epsilon $ where $\epsilon \sim \mathcal{N}(0, I)$.
Thus the encoder and decoder are trained to maximize the following objective function:
\begin{equation}\label{eq:vae_loss}
\begin{aligned}
    \mathcal{L}(\theta, \phi; x^{(i)})
    = -D_{KL}(q_{\phi}(z|x^{(i)}) || p_{\theta}(z)) + log~p_{\phi} (x^{(i)} | z^{(i)})
\end{aligned}
\end{equation}
For modelling images, both the encoder and decoder of a VAE consist of CNNs.

Higgins et al. \cite{higgins2016beta} introduce the Beta-VAE to better learn independent latent factors. The authors propose to augment the vanilla VAE loss in \Cref{eq:vae_loss} by weighing the KL term with a hyperparameter $\beta$. Choosing $\beta > $  1 enables the model to learn a more efficient latent representation of the data with better disentangled dimensions. 

\subsection{Hyperparameter Tuning of Beta-VAE}
Following Higgins et al. \cite{higgins2016beta}, we tune the hyperparameters $dim(z)$ and $\beta$ of the Beta-VAE for every dataset. To achieve maximum disentanglement, the number of latent dimensions should match the number of factors of variation in a dataset. This number is usually not known a priori. Choosing a latent space of too high dimensionality leads to a lot of uninformative dimensions with low variance. A latent space of too few latent dimensions, on the other hand, leads to entangled representations of features in the latent space. Therefore, it is necessary to identify the optimal number of latent dimensions to achieve maximally disentangled factors, ideally one in each dimension. 

To obtain the best combination of $\beta$ and $dim(z)$ for a given dataset, we first train a Beta-VAE with $\beta=1$ and $dim(z)=32$ on all datasets. We then compare the variance of the distribution in each dimension to that of a Gaussian prior. In the presence of dimensions with relatively low variance, we reduce the number of latent dimensions.
Once we find a Beta-VAE with consistently informative dimensions, i.e. with the variance comparable to the prior, we fix $dim(z)$ and focus on fine-tuning $\beta$.
As outlined by Higgings et al. \cite{higgins2016beta}, for relatively low values of $\beta$, the VAE learns an entangled latent representation since the capacity in the latent space is too high. On the other hand, for relatively high values of $\beta$, the capacity in the latent space becomes too low. The VAE performs a low-rank projection of the true data generative factors and again learns an entangled latent representation. This renders some of the latent dimensions uninformative. 

We find the highest possible $\beta$ for which all dimensions of the VAE remain informative with a variance close to the prior. In line with the findings of Higgings et al. \cite{higgins2016beta}, $\beta > 1$ is required for all datasets to achieve good disentanglement (see~\cref{tab: hyperparameter tuning}).

\begin{table}[h!]
\centering
\caption{Beta-VAE hyperparameters}
\label{tab: hyperparameter tuning}
\begin{tabular}{@{}lcc@{}}
\toprule
Dataset & $dim(z)$ & $\beta$\\
\midrule
ASVspoof & 32 & 1.25\\
CelebA & 32 & 10.0\\
Colored MNIST & 32 & 2.5\\
COVID-19 & 32 & 1.5\\
Lemon Quality & 10 & 3.0\\
Waterbirds & 32 & 1.75\\
\bottomrule
\end{tabular}
\end{table}

\subsection{Visualizing Images for Extreme $z_j$}
To validate the meaning attributed to $z_j$, we propose to additionally compute the embeddings $z_j^{(i)}$ for all the instances $x^{(i)}$ in the dataset, and identify those instances which minimize or maximize $z_j$.
Particularly, we perform $\arg\operatorname{sort}_{i} (\{z_j^{(i)}\}_{i=1}^N)$ for every dimension $j$ in the latent space and display the input images $x^{(i)}$ corresponding to the first $l$ and the last $l$ indices of the sorted values.
\Cref{fig:visualization_existing_data} depicts $l = 27$ images of the \emph{Lemon Quality} dataset with minimum and maximum values in latent dimension $2$. 

\begin{figure}
\begin{tabular}{cc}
\includegraphics[width=0.47\textwidth]{imgs/lemon_quality_images_min_dim=2.png}&
\includegraphics[width=0.47\textwidth]{imgs/lemon_quality_images_max_dim=2.png}\\
(a) Images from the \emph{Lemon Quality} & (b) Images from the \emph{Lemon Quality} \\
dataset which minimize $z_2$. & dataset which maximize $z_2$.\\
\end{tabular}
\caption{Real-world images from the training dataset that correspond to the minimum and maximum encoded values in latent dimension $2$. While bad-quality lemons are mostly captured as close-up shots, good-quality lemons are photographed from a distance.}
\label{fig:visualization_existing_data}
\end{figure}

\section{Evaluation}

\subsection{Results}

We provide illustrations of the latent space traversal for the \emph{ASVSpoof} (see \Cref{fig:asvspoof_results}) and \emph{Colored MNIST} (see \Cref{fig:cmnist_results}) datasets.


\begin{figure}[h]
    \centering
    \includegraphics[width=0.7\textwidth]{imgs/cmnist5_latentdim32_lr0.001_kld2.5_latent_traversal_dim=15.jpg}
    \caption{Evaluation of our method on the \emph{Colored MNIST} dataset.  The spurious correlation between colors and digits becomes evident with latent space traversal. The minimum values encode the colors red (left) while the maximum values encode the color blue (right).}
    \label{fig:cmnist_results}
\end{figure}

\begin{figure}[h]
    \centering
    \includegraphics[width=0.7\textwidth]{imgs/asv_kld1.25_cls0_lr0.001_dim32_latent_traversal_dim=30.jpg}
    \caption{Evaluation of our method on the \emph{ASVspoof} dataset.
    Latent traversal reaffirms the known spurious correlation between the leading silence in the audio and the target class. Leading silence (left) in the spectrogram is an indicator of benign audio samples while no leading silence (right) is common in spoofs.}
    \label{fig:asvspoof_results}
\end{figure}

\subsection{Comparison}
Heatmaps for the \emph{Lemon Quality} dataset are illustrated in \Cref{fig:lemon quality heatmaps}.

\begin{figure}[h]
\begin{tabular}{cc}
\includegraphics[width=0.47\textwidth]{imgs/lemon_heatmap_class0.png}&
\includegraphics[width=0.47\textwidth]{imgs/lemon_heatmap_class1.png}\\
(a) Class `good quality' &(b) Class `bad quality'\\
\end{tabular}
\caption{Heatmaps for images from the \emph{Lemon Quality} dataset. For the 5 most predictive features (from top to bottom) of each class we display the heatmaps on images which yield the highest activations (from left to right). The heatmaps for images of class `good quality' reveal the background as a spatial shortcut.}
\label{fig:lemon quality heatmaps}
\end{figure}
\clearpage

\bibliography{egbib}